\documentclass{article}
\usepackage{spconf,amsmath,graphicx,float}
\usepackage{makecell}
\usepackage{stfloats}
\usepackage{xcolor}

\title{A Hierarchical Slice Attention Network for Appendicitis Classification in 3D CT Scans}
\name{Chia-Wen Huang\textsuperscript{*}, Haw Hwai\textsuperscript{\dag}, Chien-Chang Lee\textsuperscript{\dag}, Pei-Yuan Wu\textsuperscript{*}}
\address{
\textsuperscript{*}Graduate Institute of Communication Engineering, National Taiwan University \\
\textsuperscript{\dag}Department of Emergency Medicine, National Taiwan University Hospital \\
\{r11942157, peiyuanwu\}@ntu.edu.tw, \{hh4832, cclee100\}@gmail.com
}

\begin{document}
%
\maketitle

%
\begin{abstract}
Timely and accurate diagnosis of appendicitis is critical in clinical settings to prevent serious complications. While CT imaging remains the standard diagnostic tool, the growing number of cases can overwhelm radiologists, potentially causing delays. In this paper, we propose a deep learning model that leverages 3D CT scans for appendicitis classification, incorporating Slice Attention mechanisms guided by external 2D datasets to enhance small lesion detection. Additionally, we introduce a hierarchical classification framework using pre-trained 2D models to differentiate between simple and complicated appendicitis. Our approach improves AUC by 3\% for appendicitis and 5.9\% for complicated appendicitis, offering a more efficient and reliable diagnostic solution compared to previous work.
\end{abstract}


\section{Introduction}
\label{sec:intro}

Accurate and timely diagnosis of appendicitis is crucial to preventing serious conditions. Computed tomography (CT) imaging plays a key role in diagnosing appendicitis and related conditions. However, the increasing number of cases can overwhelm radiologists, impacting their diagnostic capacity. To address this, automated diagnostic tools powered by deep learning have emerged as a promising solution. Developing such tools to accurately and efficiently diagnose appendicitis has therefore become a critical research focus.

One challenge in medical imaging analysis is the limited availability of labeled 3D datasets for appendicitis classification. Public datasets such as DeepLesion \cite{yan2018deeplesion}, BRATS \cite{kazerooni2023brain}, and NIH Chest X-ray \cite{filice2020crowdsourcing} are unsuitable due to their focus on other organs or imaging modalities. While Rajpurkar et al.\ \cite{rajpurkar2020appendixnet} used the Kinetics dataset for pretraining, its lack of medical relevance limits its utility for appendicitis diagnosis.

A critical challenge in 3D CT analysis is classifying small lesions, as clear markers are often lacking. Attention mechanisms help models focus on key slices, automating region selection and improving classification accuracy. Prior works, such as Park et al. \cite{park2020convolutional, park2023comparison}, used convolutional neural networks (CNNs) with manual region of interest selection, which is time-consuming. To address this issue, Sun et al.\cite{sun2023boosting} demonstrated further improvements in breast ultrasound classification by leveraging attention-guided networks, highlighting the potential of this approach.


In addition to attention mechanisms, hierarchical classification (HC) plays an important role in distinguishing between simple and complicated appendicitis, which is critical for treatment decisions. Silla et al.\cite{silla2011survey} surveyed various approaches to handling class hierarchies, and HC has been successfully implemented in medical tasks such as lesion annotation \cite{yan2019holistic} and chest X-ray localization \cite{ouyang2020learning}. Despite its success, HC has seen limited use in appendicitis research, and applying it more broadly could significantly improve diagnostic decision-making.

Although attention mechanisms and hierarchical classification have shown promise in medical imaging, their use in appendicitis diagnosis is limited. Studies like Rajpurkar et al. \cite{rajpurkar2020appendixnet}, Park et al. \cite{park2020convolutional} and Park et al. \cite{park2023comparison} rely on non-specialized pretraining or manual ROI selection, limiting its scalability. To address this, we leverage the RadImageNet dataset \cite{doi:10.1148/ryai.210315} for pretraining and automate slice selection, inspired by Sun et al. \cite{sun2023boosting}. By integrating external 2D datasets and pre-trained models, we overcome the scarcity of labeled 3D data, offering a more scalable solution for appendicitis diagnosis.

This paper makes the following contributions:
\begin{enumerate}
    \item We propose a deep learning model for appendicitis classification using 3D CT scans, incorporating external 2D datasets with slice-level annotations to automate key slice identification and improves accuracy.
    \item We introduce a hierarchical classification framework that differentiates between simple and complicated appendicitis, providing more detailed diagnostic insights to support clinical decision-making.
    \item We demonstrate that integrating pre-trained 2D models into 3D CT tasks improves classification accuracy for both appendicitis and complicated appendicitis.
\end{enumerate}

\begin{figure*}[htbp]
	\centering
    \includegraphics[scale=0.34]{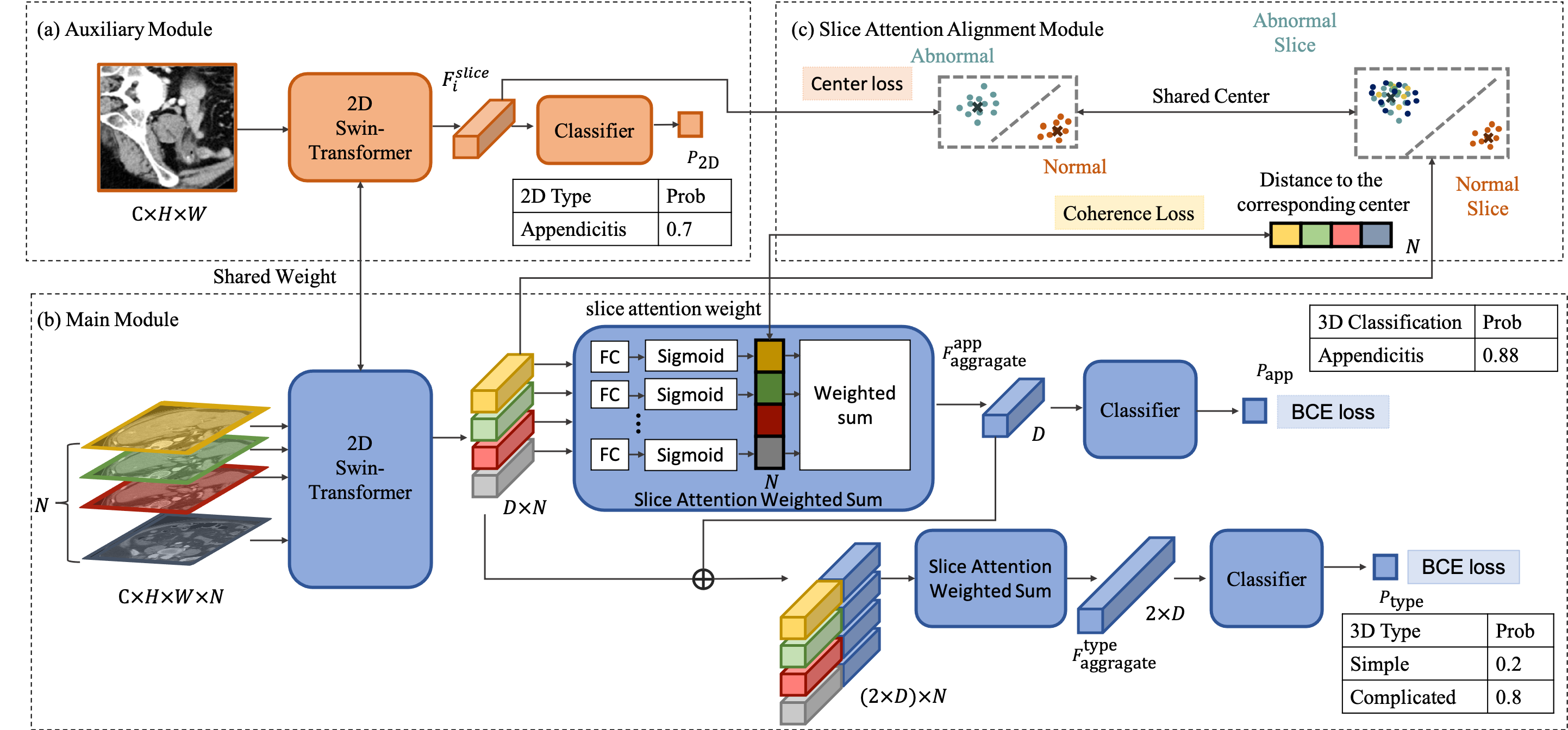}
	\caption{Overview of the proposed model. The framework includes: (a) the Auxiliary Module, which processes external 2D data with a 2D Swin Transformer; (b) the Main Module, which processes 2D slices from 3D CT scans with shared weights; and (c) the Slice Attention Alignment Module, which integrates 2D information by aligning 3D slice features with learned 2D centers using coherence loss.}
	\label{fig:overview}
\end{figure*}

\section{METHODS}
Our framework classifies 3D CT images for appendicitis and complicated appendicitis by integrating both 2D and 3D information, as shown in Fig. \ref{fig:overview}.

\subsection{Main Module: Hierarchical Classification Framework}
We adopt a hierarchical classification (HC) approach, following the Local Classifier per Level method outlined in \cite{silla2011survey}. The framework first classifies appendicitis (Appendicitis Classification) and then classifies it as either simple or complicated (Type Classification).

\textbf{Appendicitis Classification:}  
We use a 2D Swin Transformer (hereafter referred to as Swin) \cite{liu2021swin} to extract features from each CT slice. In our setup, each 3D CT scan consists of \( N \) 2D slices (in this work \( N = 64 \)). These slices are processed using a Swin model pretrained on RadImageNet \cite{doi:10.1148/ryai.210315}. 

For the task of appendicitis classification, the features from each slice are passed through an attention mechanism, where the most relevant slices for the classification task are assigned higher weights. For each slice, the attention weight 
\(
w_i^{\text{app}} = \sigma(\text{FC}(F_i^{\text{app}}))
\) is computed for appendicitis classification. The overall representation for appendicitis classification is the weighted sum of the slice features:
\begin{equation}
F_{\text{aggregate}}^{\text{app}} = \sum_{i=1}^{N} w_i^{\text{app}} \cdot F_i^{\text{app}}.
\end{equation}

The aggregated feature representation \( F_{\text{aggregate}}^{\text{app}} \) is passed to a classifier to predict whether appendicitis is present, resulting in the prediction \( P_{\text{app}}  \in \left[ 0,1 \right] \).

\textbf{Type Classification:}
The model further classifies appendicitis as either simple or complicated by concatenating the appendicitis feature representation \( F_{\text{aggregate}}^{\text{app}} \) with each slice's feature representation \( F_i^{\text{app}} \), providing both global and localized information for the type classification task.

\begin{equation}
F_{\text{i}}^{\text{type}} = \text{concat}(F_{\text{aggregate}}^{\text{app}}, F_i^{\text{app}}).
\end{equation}

The attention weights \( w_i^{\text{type}} \) for type classification are computed similarly to the appendicitis attention weights:
\(
w_i^{\text{type}} = \sigma(\text{FC}(F_i^{\text{type}}))
\). The aggregated representation for type classification is then computed as:
\begin{equation}
F_{\text{aggregate}}^{\text{type}} = \sum_{i=1}^{N} w_i^{\text{type}} \cdot F_i^{\text{type}}.
\end{equation}

Finally, the type classification is performed by passing the aggregated feature representation \( F_{\text{aggregate}}^{\text{type}} \) through a classifier, resulting in the prediction \( P_{\text{type}}  \in \left[ 0,1 \right] \).

\textbf{Loss for Hierarchical Classification}  
The HC framework employs two binary cross-entropy losses: \(\mathcal{L}_{\text{app}}\) for appendicitis classification and \(\mathcal{L}_{\text{type}}\) for type classification.

\subsection{Auxiliary Module: 2D Classification}
Inspired by Sun et al. \cite{sun2023boosting}, this module, along with the Slice Attention Alignment Module (Section \ref{sec:2-3}), leverages attention mechanisms to align 2D and 3D information for more accurate classification. The Auxiliary Module processes external 2D data labeled as either appendicitis or normal, using the same Swin as the Main Module to maintain consistency in feature extraction. By incorporating Center Loss, this module facilitates the learning of class centers for appendicitis-positive and normal slices, refining the model's attention on relevant features.

\textbf{Center Loss for Learning Class Centers:}  
Center Loss encourages 2D slices to cluster around their class centers \( C_{\text{positive}} \) (for appendicitis-positive slices) and \( C_{\text{negative}} \) (for normal slices), with each slice represented by its feature \( F^{\text{slice}}_i \) extracted by the Swin. The loss function aims to minimize the distance between each slice feature \( F^{\text{slice}}_i \) and its corresponding class center:
\begin{equation}
\begin{aligned}
\mathcal{L}_{\text{center}} = &\frac{1}{N} \sum_{i=1}^{N} \left( y_i ||F^{\text{slice}}_i - C_{\text{positive}}||_2^2 \right. \\
&\left. + (1 - y_i) ||F^{\text{slice}}_i - C_{\text{negative}}||_2^2 \right),
\end{aligned}
\end{equation}
where \( y_i \in \{ 0,1 \}\) is the ground truth label for slice \( i \).

\textbf{Loss for 2D Classification}  
The loss for 2D classification, \( \mathcal{L}_{\text{2D}} \), is computed based on the predicted probability \( P_{\text{2D}} \) of appendicitis and the corresponding ground truth label \( y_{\text{2D}} \).

\subsection{Slice Attention Alignment Module: Coherence Between 2D and 3D Slices} \label{sec:2-3}
The Slice Attention Alignment Module aligns the attention weights between the 2D and 3D slices using Coherence Loss, as described in \cite{sun2023boosting}. The learned 2D class centers from the Auxiliary Module are used to refine the attention mechanism in the 3D task. The distance between the 3D slice features \( F_i^{\text{app}} \) and the learned 2D class centers is used to calculate attention weights:
\(
d_i = ||F_i^{\text{app}} - C_{\text{positive}}||_2
\)
The inverse distance weight is then computed as:
\(
w_i^{\text{align}} = 1 - \frac{d_i}{\max(d) + \epsilon}
\).

The Coherence Loss minimizes the difference between the attention weights obtained from 3D classification and those derived from the 2D learned centers:
\begin{equation}
\mathcal{L}_{\text{coherence}} = \sum_{i=1}^{N} ||w_i^{\text{app}} - w_i^{\text{align}}||_2^2.
\end{equation}

\subsection{Total Loss Function}
The total loss function combines the losses from both the 2D and 3D tasks:
\begin{equation}
\mathcal{L}_{\text{total}} = \alpha \cdot \mathcal{L}_{\text{app}} + \beta \cdot \mathcal{L}_{\text{type}} + \gamma \cdot \mathcal{L}_{\text{center}} + \delta \cdot \mathcal{L}_{\text{coherence}} + \lambda \cdot \mathcal{L}_{\text{2D}}.
\end{equation}


\section{Experiments}
\label{sec:EXPERIMENTS}

\subsection{Dataset and Model Implementation}

\subsubsection{Dataset Overview}
This study utilized appendicitis-related datasets from two hospitals: NTUH (National Taiwan University Hospital) and FEMH (Far Eastern Memorial Hospital). Both datasets include cases of patients who underwent contrast-enhanced CT scans for suspected appendicitis. Pregnant patients, pediatric patients, and individuals with HIV were excluded.

\textbf{NTUH Dataset:}  
The NTUH dataset includes cases from patients presenting with lower abdominal or right lower quadrant pain. It comprised 2,157 cases, including 867 appendicitis-positive cases, 134 with a perforated appendix, and 82 with periappendiceal abscesses.

\textbf{FEMH Dataset:}  
The FEMH dataset included 1,056 cases, with 908 appendicitis-positive cases, 143 with a perforated appendix, and 77 with periappendiceal abscesses.

To enhance learning, RadImageNet \cite{doi:10.1148/ryai.210315}, which includes over 100,000 abdominal CT images, was used for pretraining. Additionally, AOCR2024 \cite{aocr2024}, which provides labels for each slice (unlike NTUH and FEMH, which only have whole 3D labels), was used for auxiliary tasks. Since AOCR2024 was also sourced from FEMH, steps were taken to ensure no overlap between the FEMH dataset used in this study and AOCR2024. Any overlapping cases were excluded from AOCR2024 to maintain the integrity of our FEMH test set.

Both datasets were split into 70\% training, 15\% validation, and 15\% testing subsets. Only axial CT views were used, with labeling based on original radiology reports reviewed by experts. In both datasets, perforation of the appendix and periappendiceal abscesses were classified as complicated appendicitis.

\subsubsection{Implementation Details}
The model was trained using the SGD optimizer with a momentum of 0.9 and a weight decay of 0.001. The learning rate was initially set to $5 \times 10^{-5}$ and adjusted using a learning rate scheduler based on the validation AUC for appendicitis. Training was performed with a batch size of 2 on an NVIDIA GeForce RTX 4090 GPU with 24 GB of graphics memory. To standardize the input data, each patient's CT scan---varying in the number of slices---was centered around a fixed index, selecting 64 slices in total. The 2D slices from the right lower abdomen were resized to $224 \times 224$ pixels. DICOM files were processed with fixed window center and width settings. Thresholds were set to maximize sensitivity for both simple and complicated appendicitis. These thresholds were adjusted to achieve a sensitivity of 0.9 for appendicitis and 0.8 for complicated appendicitis. These values were chosen to balance classification accuracy and minimize false negatives, given the clinical importance of avoiding missed diagnoses.

\begin{table}[t]
	\centering
    \caption{Internal validation comparison of different models.}
    \begin{tabular}{|c|c|c|c|c|}
        \hline
        Model & AUC & ACC & Sens. & Spec. \\
		\hline
  		\multicolumn{5}{|c|}{Appendicitis} \\
        \hline
        3D DenseNet  & 0.7901 & 0.6894 & \textbf{0.9023} & 0.4346 \\
  		\hline
        AppendixNet \cite{rajpurkar2020appendixnet} & 0.8585 & 0.7574 & \textbf{0.9023} & 0.5841 \\
        \hline
        \makecell[c]{Ours} & \textbf{0.8887} & \textbf{0.8064} & \textbf{0.9023} & \textbf{0.6916} \\
		\hline
        \multicolumn{5}{|c|}{Complicated Appendicitis} \\
        \hline
        3D DenseNet & 0.7846 & 0.5893 & \textbf{0.8205} & 0.5684 \\
        \hline
        AppendixNet \cite{rajpurkar2020appendixnet} & 0.8244 & 0.6404 & \textbf{0.8205} & 0.6241 \\
        \hline
        \makecell[c]{Ours} & \textbf{0.8833} & \textbf{0.8362} & \textbf{0.8205} & \textbf{0.8376} \\
		\hline
     \end{tabular}
     \label{table:NTUH_FEMH}
\end{table}


\begin{table}[t]
	\centering
    \caption{External validation comparison of different models.}
	\begin{tabular}{|c|c|c|c|c|}
        \hline
        Model & AUC & ACC & Sens. & Spec. \\
		\hline
  		\multicolumn{5}{|c|}{Appendicitis} \\
  		\hline
        3D DenseNet  & 0.4496 & 0.4190 & 0.896 & 0.1050 \\
        \hline
        AppendixNet \cite{rajpurkar2020appendixnet} & 0.7044 & 0.4921 & \textbf{0.904} & 0.2211 \\
        \hline
        \makecell[c]{Ours} & \textbf{0.7413} & \textbf{0.5556} &  \textbf{0.904} & \textbf{0.3263} \\
		\hline
        \multicolumn{5}{|c|}{Complicated Appendicitis} \\
        \hline
        3D DenseNet  & 0.5772 & 0.3651 & \textbf{0.8421} & 0.3345 \\
        \hline
        AppendixNet \cite{rajpurkar2020appendixnet} & 0.7601 & 0.5619 & \textbf{0.8421} & 0.5439 \\
        \hline
        \makecell[c]{Ours} & \textbf{0.8803} & \textbf{0.8317} & \textbf{0.8421} & \textbf{0.8310} \\
		\hline
     \end{tabular}
     \label{table:External}
\end{table}

\begin{table}[t]
    \centering
    \caption{Ablation Study}
    \begin{tabular}{|c|c|c|c|c|}
        \hline
        Model & AUC & ACC & Sensitivity & Specificity \\
        \hline
        \multicolumn{5}{|c|}{Appendicitis} \\
        \hline
        Base  & 0.8674 & 0.7787 & \textbf{0.9023} & 0.6308 \\
        \hline
        \makecell[c]{Base \\ + hierarchy} & 0.8847 & 0.7936 & \textbf{0.9023} & 0.6636 \\
        \hline
        \makecell[c]{Base \\ + hierarchy \\ + 2D} & \textbf{0.8887} & \textbf{0.8064} & \textbf{0.9023} & \textbf{0.6916} \\
        \hline
        \multicolumn{5}{|c|}{Complicated Appendicitis} \\
        \hline
        Base & 0.8695 & 0.7851 & \textbf{0.8205} & 0.7819 \\
        \hline
        \makecell[c]{Base \\ + hierarchy} & 0.8804 & 0.8000 & \textbf{0.8205} & 0.7981 \\
        \hline
        \makecell[c]{Base \\ + hierarchy \\ + 2D} & \textbf{0.8833} & \textbf{0.8362} & \textbf{0.8205} & \textbf{0.8376} \\
        \hline
    \end{tabular}
    \label{table:Ablation_Study}
\end{table}

\subsection{Model Performance}
We conducted both internal and external validations to evaluate model performance across different clinical settings. Internal validation involved mixed training with the NTUH and FEMH datasets, while external validation was performed by training on FEMH and testing on NTUH to assess generalization.

In this study, we compared our model to AppendixNet \cite{rajpurkar2020appendixnet}, which is specifically designed for appendicitis diagnosis, and 3D DenseNet, a general-purpose model for 3D imaging. The methods proposed by Park et al. \cite{park2020convolutional}, \cite{park2023comparison} were excluded from this comparison due to their requirement for manual ROI selection, which is incompatible with our automated approach.

Tables \ref{table:NTUH_FEMH} and \ref{table:External} present the internal and external validation results, respectively. The metrics compared include AUC, accuracy (ACC), sensitivity (Sens.), and specificity (Spec.). Our model consistently outperformed both 3D DenseNet and AppendixNet in internal and external validations, achieving higher AUC and ACC values for both appendicitis and complicated appendicitis. Sensitivity remained constant due to fixed thresholds, but the model showed significant improvements in specificity, reducing false positives.

\subsection{Ablation Study}
The ablation study, conducted using the same dataset as in the internal validation, is presented in Table \ref{table:Ablation_Study}. The \textbf{base model}, a 2D Swin Transformer pretrained on RadImageNet, processes each slice independently and uses a slice-weighted sum for classification. Adding our hierarchical classification module (\textbf{base+hierarchy}) improved AUC and specificity, while incorporating 2D slice analysis with external datasets (\textbf{base+hierarchy+2D}) enhanced attention to key slices, improving the distinction between simple and complicated appendicitis.

\subsection{Discussion}
The results in Tables \ref{table:NTUH_FEMH} and \ref{table:External} confirm that high-quality pretrained data improves model robustness, as demonstrated by better performance. External 2D datasets with slice-level labels further enhance key slice learning, even for datasets without slice annotations, improving appendicitis classification. The ablation study in Table \ref{table:Ablation_Study} highlights the importance of hierarchical classification, which may aid in distinguishing complicated appendicitis from other conditions, such as periappendiceal or abdominal abscesses. Notably, complicated appendicitis, such as an appendiceal abscess, can only occur in the presence of appendicitis; abscesses unrelated to the appendix are not considered complicated appendicitis. Lastly, the Slice Attention Alignment Module enabled the model to focus on key slices in CT scans with numerous slices.

\section{CONCLUSION}
\label{sec:CONCLUSION}

In conclusion, we propose a 2D hierarchical deep learning model for classifying appendicitis and complicated appendicitis from 3D CT scans. Leveraging Slice Attention and external 2D datasets enhances small lesion classification, and integrating pre-trained 2D models improves accuracy and robustness. Our approach effectively distinguishes between simple and complicated appendicitis, demonstrating the value of combining 2D and 3D information with attention mechanisms for clinical applications.


\section{COMPLIANCE WITH ETHICAL STANDARDS}
\label{sec:ethical_standards}
This retrospective study was conducted following the Declaration of Helsinki and was approved by the Research Ethics Committee of National Taiwan University Hospital. Due to the retrospective nature of the study, informed consent was waived. A focused database extraction ensured that only relevant data were retrieved and analyzed. All personal information was masked to maintain patient confidentiality.
IRB Approval Number: 202201049RINB




\bibliographystyle{IEEEbib}
\bibliography{refs}

\end{document}